\pdfoutput=1
\documentclass{article}
\usepackage{arxiv}
\usepackage[T1]{fontenc}
\usepackage{authblk}
\newcommand{\figcolwidth}{0.6\textwidth}
\newenvironment{IEEEkeywords}{\par\noindent\textbf{\emph{Keywords}}\enspace}{\par}

\usepackage{cite}
\usepackage{amsmath,amssymb,amsfonts}
\usepackage{graphicx}
\usepackage{booktabs}
\usepackage{textcomp}
\usepackage{xcolor}
\usepackage{url}
\usepackage[hidelinks]{hyperref}
\makeatletter
\providecommand{\bstctlcite}[1]{\if@filesw\immediate\write\@auxout{\string\citation{#1}}\fi}
\makeatother

\begin{document}
\bstctlcite{IEEEexample:BSTcontrol}

\title{SARA: An Auditor-in-the-Loop LLM Recommender for Sustainability Reports}
\date{}
\renewcommand\Authfont{\bfseries}
\author[1,2,$\dagger$]{Lars Hillebrand\thanks{\texttt{lars.patrick.hillebrand@iais.fraunhofer.de}.
ORCID: L. Hillebrand 0000-0002-5496-4177; T. B. F. de Oliveira 0000-0002-4493-8012.}}
\author[1,$\dagger$]{Thiago B. F. de Oliveira}
\author[3]{Tim Dilmaghani}
\author[3]{Mohamed Khaled}
\author[1]{Writwick Das}
\author[1,2,4]{Rafet Sifa}
\affil[1]{Fraunhofer IAIS, Sankt Augustin, Germany}
\affil[2]{Lamarr Institute for Machine Learning and Artificial Intelligence, Bonn, Germany}
\affil[3]{PwC GmbH WPG, Frankfurt, Germany}
\affil[4]{University of Bonn, Bonn, Germany}
\affil[$\dagger$]{Both authors contributed equally to this research.}

\maketitle
{\renewcommand{\thefootnote}{}\footnotetext{\copyright~2026 IEEE. Personal use of this material is permitted. Permission from IEEE must be obtained for all other uses, in any current or future media, including reprinting/republishing this material for advertising or promotional purposes, creating new collective works, for resale or redistribution to servers or lists, or reuse of any copyrighted component of this work in other works. Accepted as a poster paper at the 2026 IEEE International Conference on Big Data (IEEE BigData 2026).}}

\begin{abstract}
Auditing corporate sustainability reports requires assurance professionals to
map a company-specific checklist drawn from roughly 1,700 regulatory
requirements onto the relevant passages of lengthy, repeatedly revised draft
reports and to judge compliance. We present SARA (Sustainability
Reporting Assistant), a human-in-the-loop recommender system delivered as a
Microsoft Word add-in that recommends, for each requirement, the passages
addressing it and predicts a compliance verdict while keeping the
auditor in full control. SARA casts auditing as content-based recommendation
over a fixed regulatory taxonomy and makes per-requirement large language model
(LLM) matching cheaper, faster, and more accurate through
topic-prediction filtering: on expert-annotated reports, it reduces matching cost
about sixfold and runtime by 42\% versus feeding the full report to the LLM. It
is in production use across PwC Germany's sustainability-assurance practice.
\end{abstract}


\begin{figure*}[!t]
  \centering
  \includegraphics[width=\textwidth]{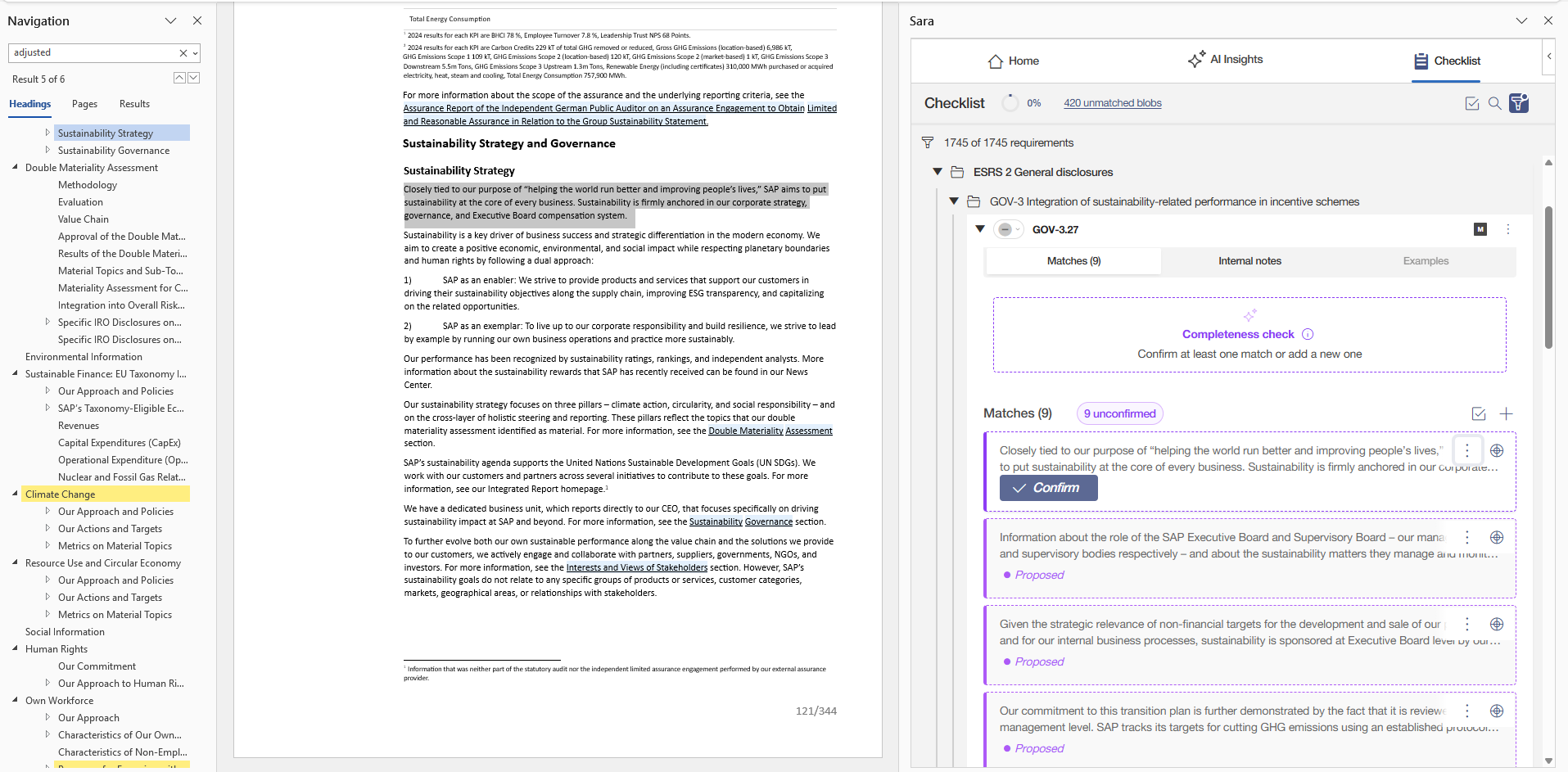}
  \caption{SARA in Microsoft Word: the task pane (right) shows the audit
  checklist with one requirement expanded and its recommended passages; the
  passage of the first recommendation is highlighted in the document (left).
  The report is publicly available and used for illustration only.}
  \label{fig:screenshot}
\end{figure*}

\section{Introduction}
The EU's Corporate Sustainability Reporting Directive (CSRD)~\cite{eu2022csrd}
has sharply increased the volume and complexity of corporate sustainability
disclosures: its reporting standards define roughly 1{,}700 regulatory
requirements, and a materiality assessment determines the applicable subset per
company. Auditing a report requires mapping each applicable requirement to the
passages that address it and judging whether it is fulfilled. This is
predominantly manual, repeated for every new report version, and complicated by
report heterogeneity: disclosures span narrative text, tables, and lists. Large
language models (LLMs) can automate the matching, but prompting them
with the whole report for each of hundreds of applicable requirements is slow
and costly.

We present \textbf{SARA} (Sustainability Reporting Assistant), a
human-in-the-loop recommender system implemented as a Microsoft Word add-in
(Fig.~\ref{fig:screenshot}). SARA recommends the passages addressing each
requirement and predicts a compliance verdict. Beyond recommendation, it
structures the entire audit workflow inside Word, where audits are carried out,
removing tool switching and replacing a manual, spreadsheet-based process. We make three contributions:
\begin{enumerate}
  \item a \emph{content-based, zero-shot recommender over a fixed regulatory
    taxonomy}, recommending a report's passages for each of roughly
    1{,}700 regulatory requirements;
  \item \emph{topic-prediction filtering} that makes per-requirement LLM matching
    cheaper, faster, and more accurate; and
  \item \emph{expert-in-the-loop review} with a persistent, versioned
    audit trail: auditor decisions override the model, recompute verdicts, and
    persist across report versions.
\end{enumerate}

\section{Related Work}
SARA continues a line of work on machine-learning-assisted auditing:
ALI~\cite{sifa2019ali} framed requirement-to-passage mapping as content-based
recommendation~\cite{ricci2011handbook}, a zero-shot LLM stage
filtering its recommendations improved matching for financial
audits~\cite{hillebrand2023zeroshot}, sustain.AI~\cite{hillebrand2023sustainai}
extended it to sustainability reports with a fine-tuned BERT classifier,
pointer-guided pre-training~\cite{hillebrand2024pointer} infused paragraph-level
context into document representations, contradiction
detection~\cite{deusser2023contradiction} checks financial reports for
consistency, and verification regimes~\cite{sifa2026verification} delimit
which audit checks LLMs can take over and where the auditor's judgment stays
essential. SARA replaces the trained classifier with
an LLM matching pipeline; instead of retrieval-augmented
generation~\cite{lewis2020rag} over dense
retrieval~\cite{karpukhin2020dpr,reimers2019sbert}, it selects candidates by
predicted sub-topic.
LLMs are competitive zero-shot recommenders when preferences are expressed in
natural language~\cite{sanner2023llmrec} and effective list-wise rankers with a
favorable cost--quality trade-off~\cite{dai2023chatgptrec}; SARA
operationalizes exactly this list-wise matching. LLMs also accelerate
regulatory-text analysis, extracting requirements from legal
sources~\cite{abualhaija2025llmregulatory} and structured disclosures from ESG
reports~\cite{zou2025esgreveal}, while conversational recommenders refine
relevance through user feedback~\cite{jannach2021conversational}. SARA differs by
treating each requirement as a query over a fixed, expert-curated taxonomy
rather than an open item corpus, in a high-stakes setting where the auditor
remains the final arbiter.

\section{System}
Audits are conducted almost entirely in Word: drafts arrive as \texttt{.docx}
files, auditors annotate them, and clients return revised versions, making the
document the single source of truth. SARA therefore runs as an
add-in built on the Office.js API. It reads content and layout directly from
the Word object model, and linking a passage to a requirement takes a text
selection and one click.

\subsection{Topic Prediction and Matching}
\label{sec:matching}
The checklist is a two-level taxonomy: \emph{topics} (e.g.\ climate change,
own workforce) contain \emph{sub-topics} (e.g.\ transition plan for climate
change mitigation)\footnote{Topics correspond to the European Sustainability
Reporting Standards (ESRS, e.g.\ E1 Climate change); sub-topics to their
disclosure requirements, not to the sub-topics defined in ESRS.}, and the
natural-language requirements sit as leaves under
the sub-topics. A first stage, \emph{topic prediction}, places the report in the
same taxonomy: an LLM assigns every passage (paragraph, table, or list item)
one or more topics and sub-topics, since a passage can address requirements
from several sub-topics. Requirements and passages thus share one
coordinate system.

Matching then proceeds requirement by requirement. SARA restricts the candidates
to passages sharing the requirement's sub-topic and prompts an LLM to select those
addressing the requirement and to assign a provisional verdict of
\textit{Fulfilled}, \textit{Partially Fulfilled}, or \textit{Not Fulfilled} in
the same call. The filter is conservative: it keeps passages whose topic prediction is
uncertain and falls back to the full report if no candidates remain. Matching against a small, sub-topic-relevant slice
rather than the whole report cuts input tokens and reduces the
needle-in-a-haystack effect of long-context matching
(Section~\ref{sec:eval}). The same topic predictions power a peer-group benchmarking view
that contrasts the report's topic distribution with 500+ peer reports,
surfacing under- and over-reported topics.

\subsection{Auditor-in-the-Loop Review}
\begin{figure}[!t]
  \centering
  \includegraphics[width=\figcolwidth]{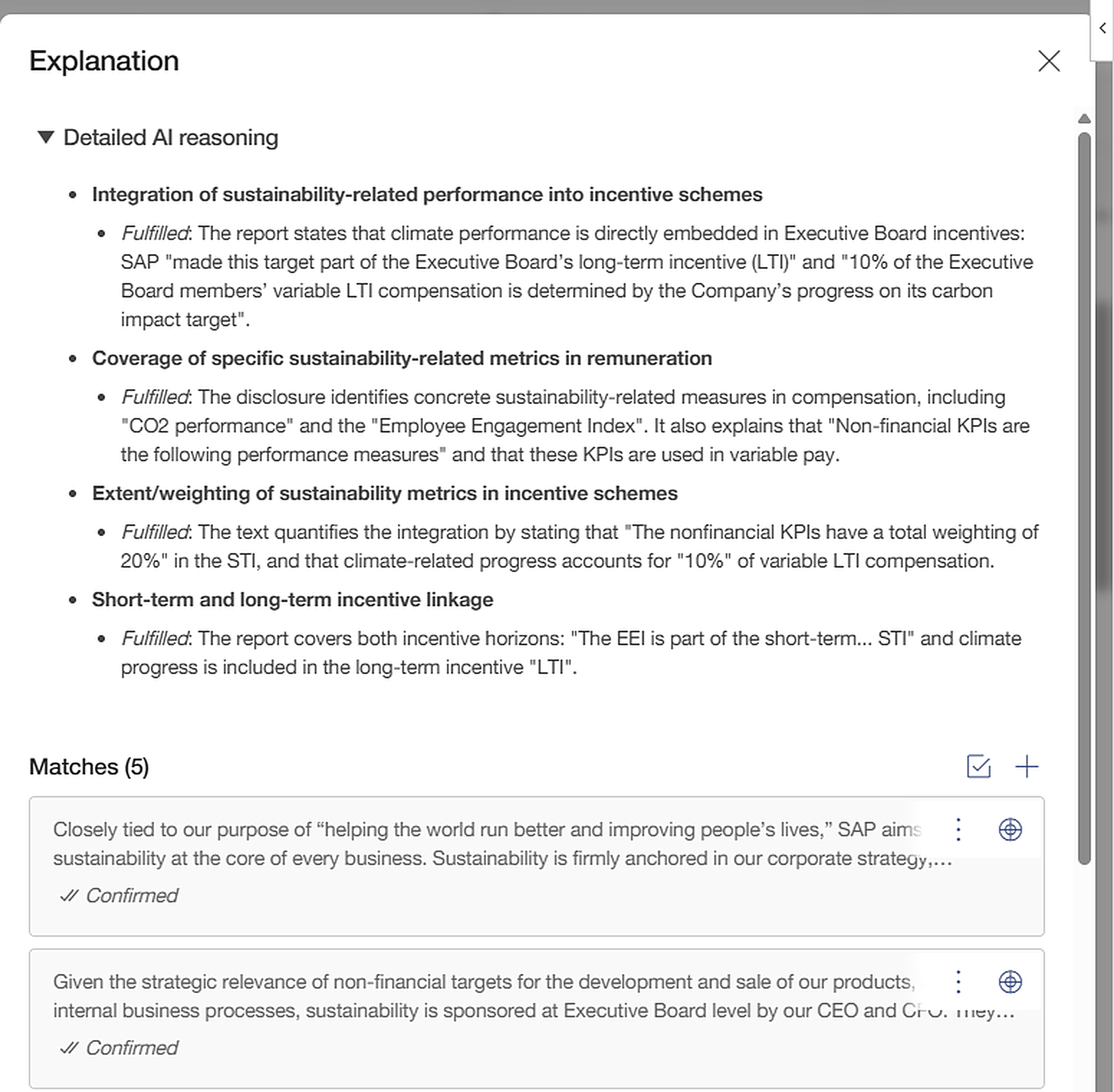}
  \caption{Verdict explanation in the task pane: per-aspect verdicts with quoted
  report text, followed by the matches and their confirmation status.}
  \label{fig:explanation}
\end{figure}

The selected passages appear in the task pane as \emph{proposed matches},
organized by topic and sub-topic. Each match links to its location: selecting it
scrolls the document to the passage and highlights it in place. The auditor
confirms or rejects proposed matches and adds manual ones by selecting text in
the document, so the auditor, not the model, makes the final relevance
decision. Confirmed matches stay anchored to their document location, and
every change to the match set recomputes the verdict over the
confirmed matches. The verdict is explained rather than asserted: SARA
decomposes the requirement into its aspects and states for each whether the
matches satisfy it, quoting the supporting report text
(Fig.~\ref{fig:explanation}). All actions are persisted, yielding an audit trail.

\subsection{Report Versions}
Audits are iterative: the client addresses the auditor's findings and submits a
revised version, typically three to six times per engagement. SARA carries recorded matches, comments, and verdicts forward to the
next version and diffs the two documents. Matches anchored to modified or deleted
passages are flagged as pending, so the auditor re-examines
only what changed and re-audit effort scales with the size of the revision, not
the report.

\section{Evaluation}
\label{sec:eval}
We compare three candidate strategies for LLM matching.
\textbf{S1 (Azure hybrid)} indexes the chunked report in Azure AI Search and
retrieves the top-$k$ chunks per requirement by hybrid vector-and-keyword
search. \textbf{S2 (full report)} passes all passages to the LLM per
requirement, exploiting long context and prompt caching. \textbf{S3
(topic-filtered)} is the deployed configuration of
Section~\ref{sec:matching}, restricting candidates to the requirement's
predicted sub-topic. All three run \texttt{gpt-5.4}.

We evaluate on $N{=}2$ expert-annotated CSRD reports (477 and 397 annotated
requirements, 874 in total; other reports lacked sufficient annotation
quality). The reports come from different companies and industries and differ
markedly in length, structure, and topics; S3 alone makes 3{,}948
requirement--passage match decisions on them. Compliance verdicts are not
annotated, so we report matching quality. We favor recall: a missed passage
costs the auditor more than an easily dismissed false positive. Precision, recall, and F1 are macro-averaged
over requirements pooled across both reports; F1 is the mean of
per-requirement F1 scores.

\begin{table}[!t]
\caption{Strategies on $N{=}2$ reports (874 annotated requirements), all
\texttt{gpt-5.4}; macro metrics over requirements; cost and time are
totals over both reports.}
\label{tab:eval}
\begin{center}
\setlength{\tabcolsep}{4.5pt}
\begin{tabular}{lccccc}
\toprule
\textbf{Strategy} & \textbf{Prec.} & \textbf{Rec.} & \textbf{F1} &
\textbf{Cost (\$)} & \textbf{Time (min)} \\
\midrule
S1 (Azure hybrid) & 42.2 & 68.0 & 45.7 & 52.9 & 10.3 \\
S2 (full report)  & 44.2 & 70.2 & 47.5 & 126.2 & 12.9 \\
S3 (topic-filtered) & \textbf{48.6} & \textbf{70.6} & \textbf{50.6} & \textbf{19.6} & \textbf{7.5} \\
\bottomrule
\end{tabular}
\end{center}
\end{table}

As Table~\ref{tab:eval} shows, S3 is the cheapest, fastest, and most accurate
strategy. Matching each requirement against only its sub-topic's
passages cuts input tokens by ${\sim}95\%$ relative to S2, making S3
${\sim}6\times$ cheaper than S2 and $2.7\times$ cheaper than S1, with 42\% less
runtime than S2 even including topic prediction. It also raises precision
and F1 over S2 by 4.4 and 3.1 points, and its slightly higher recall indicates
that the conservative filter rarely discards relevant passages.

\section{Conclusion}
SARA embeds requirement-to-passage recommendation and compliance prediction into
the auditor's Word environment and organizes the full sustainability-audit
workflow around it.
Topic-prediction filtering makes per-requirement LLM matching cheaper and
more accurate. Because SARA runs inside PwC's production environment
and is not publicly accessible, we provide a screencast at
\mbox{\url{https://s.fhg.de/sara}}. Future work will
close the feedback loop, using stored auditor decisions to refine the
recommender, and extend the evaluation to compliance verdicts once
annotated.

\section*{Acknowledgments}
This research has been partially funded by the Federal Ministry of Education
and Research of Germany and the state of North-Rhine Westphalia as part of the
Lamarr Institute for Machine Learning and Artificial Intelligence.

\bibliographystyle{IEEEtran}
\bibliography{references}

\end{document}